\documentclass[sigconf]{acmart}
\renewcommand\footnotetextcopyrightpermission[1]{} 
\makeatletter
\renewcommand\@formatdoi[1]{\ignorespaces}
\makeatother
\usepackage{graphicx}
\usepackage{caption}
\usepackage{subcaption}

\AtBeginDocument{%
  \providecommand\BibTeX{{%
    \normalfont B\kern-0.5em{\scshape i\kern-0.25em b}\kern-0.8em\TeX}}}

\acmConference[Woodstock '18]{Woodstock '18: ACM Symposium on Neural
   Gaze Detection}{June 03--05, 2018}{Woodstock, NY}
\acmBooktitle{Woodstock '18: ACM Symposium on Neural Gaze Detection,
  June 03--05, 2018, Woodstock, NY}
\acmPrice{15.00}
\acmISBN{978-1-4503-XXXX-X/18/06}

\begin{document}

\title{Performance in the Courtroom: Automated Processing and Visualization of Appeal Court Decisions in France}


\author{Paul Boniol}
\email{paul.boniol@polytechnique.edu}
\affiliation{%
  \institution{LIX,  \'Ecole Polytechnique}
  \city{Paris}
  \country{France}}
  
  \author{George Panagopoulos}
\email{george.panagopoulos@polytechnique.edu}
\affiliation{%
  \institution{LIX,  \'Ecole Polytechnique}
  \city{Paris}
  \country{France}}
  
\author{Christos Xypolopoulos}
\email{christos.xypolopoulos@polytechnique.edu}
\affiliation{%
  \institution{LIX,  \'Ecole Polytechnique}
  \city{Paris}
  \country{France}}

\author{Rajaa El Hamdani}
\email{el-hamdani@hec.fr}
\affiliation{%
  \institution{HEC Paris}
  \city{Paris}
  \country{France}}

\author{David Restrepo Amariles}
\email{restrepo-amariles@hec.fr}
\affiliation{%
  \institution{HEC Paris}
  \city{Paris}
  \country{France}}
  
\author{Michalis Vazirgiannis}
\email{mvazirg@lix.polytechnique.fr}
\orcid{0000-0001-5923-4440}
\affiliation{%
  \institution{LIX,  \'Ecole Polytechnique}
  \city{Paris}
  \country{France}
}

\renewcommand{\shortauthors}{Vazirgiannis and Restrepo, et al..}

\begin{abstract}
Artificial Intelligence techniques are already popular and important in the legal domain. We extract legal indicators from judicial judgments to decrease the asymmetry of information of the legal system and the access-to-justice gap. We use NLP methods to extract interesting entities/data from judgments to construct networks of lawyers and judgments. We propose metrics to rank lawyers based on their experience, wins/loss ratio and their importance in the network of lawyers. We also perform community detection in the  network of judgments and propose metrics to represent the difficulty of cases capitalising on communities features.
\end{abstract}




\maketitle
\pagestyle{plain}

\section{Introduction}
Recent advances in Artificial Intelligence (AI) and Natural Language Processing (NLP) allow the analysis of large numbers of legal documents in aggregate in contrast to traditional methods. A long-standing application of NLP to legal documents is information extraction and retrieval from judicial decisions. The interest in mining data from judgments can be explained by the critical role they play in the administration of justice in both common and civil law systems. The objective of our work is to analyze judgments by French courts to gain insights about the operation of the French judicial system, which could in turn help developing an interface for laypersons. As explained in \cite{ruhl2015measuring}, a legal user interface could shield the user of the legal system from the complexity of the underlying legal system. 
Ordinary people perceive the legal system as too complex \cite{oecd2016a2j}, which results in part from the asymmetry of information in the market of legal services, where ordinary people are disadvantaged in comparison with providers of legal services \cite{slaw_2016}. The asymmetry of information adds to the access-to-justice gap, such that a layperson lacks the right information and tools to choose the right lawyer at an affordable cost, and might prefer to self represent herself or refrain from filing a lawsuit. According to \cite{greacen2014market} "one of six Americans is a self-represented litigant in a newly filed case each year," however, the resolutions are in favor of litigants represented with a lawyer. Both \cite{slaw_2016, greacen2014market} suggests "the ease of access to information" is a solution to address the gap in accessing justice. Access to free basic legal information could help the user to navigate the justice system easily, understand better the legal area his problem falls into, and choose a lawyer with experience on the subject matter of the dispute.  
In our work, we extract and represent information from past judgments to increase the transparency of judicial procedures and make them more accessible to laypersons. First, we pre-process judgments by extracting relevant legal entities, such as the lawyers of each party, by using Named Entity Recognition (NER) models. Second, we analyze the win/loss rate of lawyers by building two lawyers' networks: an opposing network of lawyers and a collaborative network of lawyers. Third, we use network analysis of judgments to suggest a measure of case difficulty based on case types/communities with distinct win/lose rates.
 
\section{Related work}
Numerous research have been carried out on case-law corpora focusing on specific objectives. One of the long-time objectives is the prediction of case outcomes. One of the first approaches in this field was \cite{kort1957predicting} to manually convert a case factual elements into numerical values, compute their sum and predict a decision in favor of the petitioner if the sum is above a manually selected threshold. Recent efforts \cite{katz2014predicting, sulea2017predicting, branting2019semi, long2019automatic} have used machine learning techniques to build outcome prediction models.
Judicial judgments are rich in data, which could be used to analyze the operations of the legal system. The authors of \cite{epstein2013behavior, rachlinski2017judging} used empirical methods to understand and describe judicial decision-making. Other researchers extract information to empower legal decision-makers and legal practitioners \cite{michalopoulos2019ai, mok2019legal, howe2019legal, vacek2019litigation}.
Judicial decisions lend themselves to the use of network analysis techniques. Networks of case law have been used several times \cite{fowler2007network, derlen2014goodbye} to measure the importance of a case. The theory of graphs provides tools well adapted to analyze the complexity of case law networks; for example, \cite{tarissan2016analysing} employs a hybrid version for bipartite graphs to clarify procedural aspects of the International Criminal Court.
Judgments are expressed in natural language, therefore to scale their automatic processing, several researchers have been developing natural language processing techniques for the legal domain. Some adapt NLP techniques built for the general language to the legal language. \cite{sanchez2019sentence} build their model of sentence boundary detection (SBD) for legal documents. Researchers from the Lynx project \cite{rehm2019developing} developed a set of NLP services to extract a variety of information from legal documents: term extraction, text structure recognition, and NER. NER techniques have several applications in the legal domain. \cite{barriere2019may} improved existing NER models and used the resulting models to extract, from French judgments, entities that should be anonymized before the public release of the judgments.

\section{Data}
\subsection{Data collection}
Our dataset consists of 40,000 rtf files that were crawled through L\'egifrance \footnote{\url{www.legifrance.gouv.fr}}, a French legal publisher providing access to law codes and legal decisions. To navigate and crawl through L\'egifrance we used Selenium\footnote{https://selenium-python.readthedocs.io/}, a python framework that simulates a real web browser. For our experiments, we used a sample of cases from the court of appeal consisting of 17,215 cases. We limit our first analysis to cases from the court of appeal due to the specificity of cases from trial courts and the Court of Cassation. For future works and analysis, a sample of cases from the Court of Cassation could also be used (more than 400,000 documents available on L\'egifrance).
We decided to focus first on cases decided by civil courts and to exclude both criminal, administrative, and specialized courts. We also remove procedural judgments, such as court orders. Judgments analyzed here are solely final decisions called "arrêt de Cour d'appel."

\subsection{Data preprocessing}
Data preprocessing was the most challenging part of the project. The structure and wording of the legal documents, which vary between different courts and dockets, as well as the use of legal formal language, were challenging obstacles to conduct the text mining tasks. Below we analyze in detail how we approached each part, from segmenting the documents to extracting the persons taking part in each court case and their roles.
\subsubsection{Segmentation}
\paragraph{\textbf{Analysis of the macrostructure of cases}}
The decisions of courts of appeals in France follow an overall similar structure. First, the documents state practical information about the litigation such as dates, jurisdiction, and the different entities involved in the trial, listed in the following order:
\begin{itemize}
    \item \textbf{Appellant (appelant in French)}: The name of the party is always anonymized, for example: "Monsieur Jean X."
    \item \textbf{Appellant's counsel}: can be anonymized but always start with the keyword "Me" or "Ma\^itre," for example, "Me Jean Dupont."
    \item \textbf{Appellee (intim\'ee in French)}: this entity has the same format as the appellant.
    \item \textbf{Appellee's counsel}: Same format as the appellant's counsel.
    \item \textbf{Court Entities} (non-fixed order):
    \begin{itemize}
        \item \textbf{Judge (magistrats, conseillers in French )}: could be anonymized but always start with the keyword "Pr\'esident."
        \item \textbf{Clerk (Greffier in French)}: Can be anonymized but is always expressed near the word "Greffier."
    \end{itemize}
\end{itemize}
After listing the entities, decisions from French appeal courts continue with the debate. The debate describes all the facts and procedure leading to the appeal. It also states the different arguments brought forth by the parties, and follows with the reasoning of the court. Finally, it closes with the conclusion which states the final decision. The keywords separating the different parts vary significantly, and are sometimes absent, which makes the segmentation task complex. Keywords may vary from one appeal court to another. We use graphs to compare the structure of judgments of appeal courts in several territorial jurisdictions. Figure 1 represents the flow of cases in two different jurisdictions. Each graph is built by parsing judgments from the same jurisdiction into sentences and then linking consecutive sentences by an edge. The name of a node is the text of the sentence. When the sentence is more than five words, the name of the node is \texttt{"Long\_Text\_i\_j"} where i is the index of the case in the whole dataset, and j is the index of the sentence within the case. The size of the node is the occurrence of its text in all the judgment of the considered jurisdiction. To account for keywords that have small variations across documents, we use the Jaro similarity to identify these variations, examples are in table \ref{tab:jaro}. The Jaro similarity is a similarity measure between two strings $s_1$ and $s_2$ \cite{jaro1989advances} defined with the following formula:
\begin{align*}
    Sim_J(s_1, s_2) =
    \begin{cases}
    0 & \text{if m = 0,}\\
    \frac{1}{3}(\frac{m}{\mid s_1 \mid} + \frac{m}{\mid s_2 \mid} + \frac{m-t}{m}) & \text{otherwise}
    \end{cases}
\end{align*}
Where:
\begin{itemize}
    \item $\mid s_i\mid$ is the length of string $s_i$
    \item m is the number of same characters not further than $\lfloor \frac{max(\mid s_1 \mid, \mid s_2 \mid)}{2}\rfloor -1$
    \item t is the number of transpositions.
\end{itemize}
The Jaro similarity is used to contract nodes of similar sentences, such that if two sentences have a Jaro similarity larger than 0.8, then they are considered belonging to the same node. Therefore the big nodes are common parts from all documents, and they represent the structure of these documents.
\begin{table*}
  \caption{Examples of Jaro distance between pairs of similar sentences contracted to a same node}
  \label{tab:jaro}
  \begin{tabular}{cccc}
    \toprule
    Sentence 1&Sentence 2& Jaro distance\\
    \midrule
    faits et procedure &faits procedure & 0.86\\
    procedure et pretentions des parties & procedure et moyens des parties & 0.83\\
    moyens et pretentions des parties &pretentions et moyens des parties& 0.92\\
  \bottomrule
\end{tabular}
\end{table*}

\begin{figure}[h]
  \centering
  \begin{minipage}[b]{0.48\linewidth}
    \centering\includegraphics[width=\textwidth]{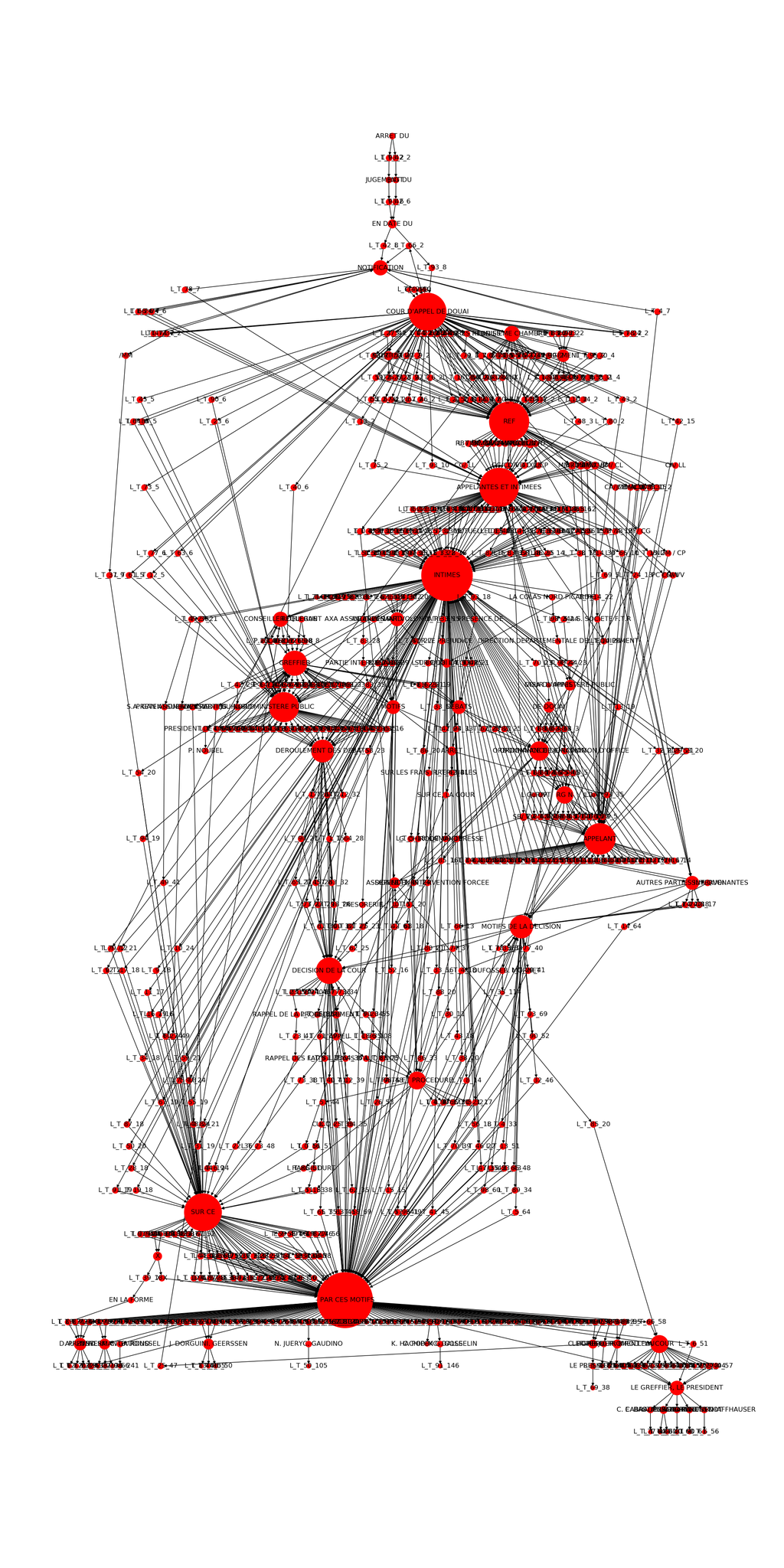}
  \end{minipage}
  \hfill
  \begin{minipage}[b]{0.48\linewidth}
    \centering\includegraphics[width=\textwidth]{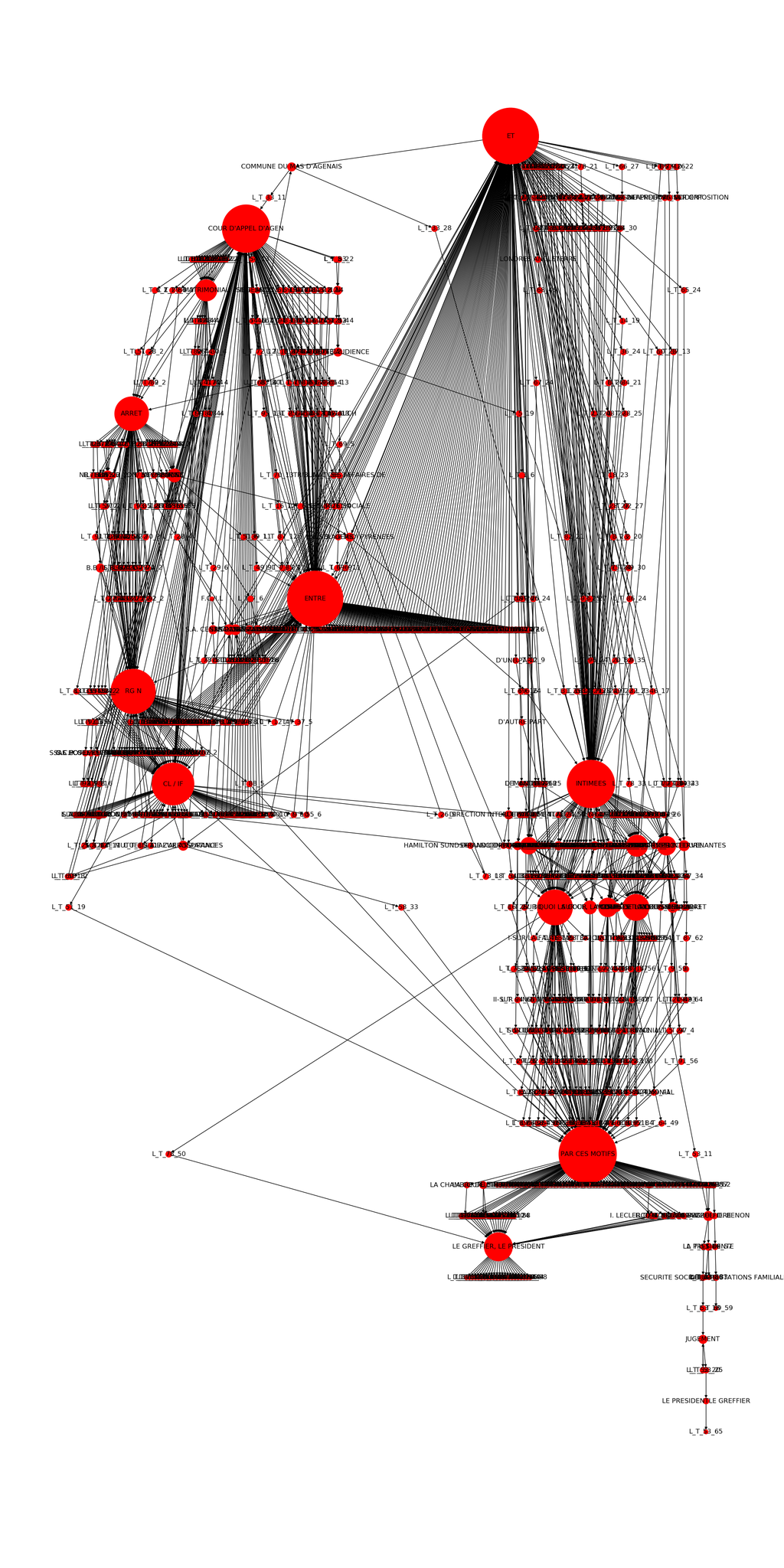}
  \end{minipage}
  \caption{Most recurrent Keywords in decisions by courts in Douai, and Agen respectively. The nodes are part of the document. When the sentence is more than five words, the name of the node is attributed to \textit{\texttt{Long\_Text\_i}} with i unique. Therefore, the big nodes are common parts from all documents, and therefore the structure.}
  \label{fig: case structure}
\end{figure}

Figure \ref{fig: case structure} points out the difference in structure and flows that documents from different jurisdictions can have. For instance, Agen will use "ENTRE" to announce the appellant, and "ET" to announce the appellee, whereas Douai will use respectively "APPELANT" and "INTIMEE." 

Nevertheless, we empirically observed that all the decisions, whatever the jurisdiction was, shared the same keyword "PAR CES MOTIFS" to announce the final decision of the court (last big node in the two flows of figure 1).

\begin{figure}[h]
  \centering
  \includegraphics[width=0.4\linewidth]{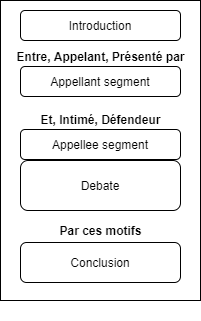}
  \caption{Segmentation of a legal case based on keywords, to facilitate entity recognition}
  \label{fig:case segments}
\end{figure}

\paragraph{\textbf{Segmentation with keywords}}
We also sought to extract entities corresponding to the lawyers defending each party. As described above, legal entities are mentioned after the practical information in a fixed order. Moreover, domain experts confirmed these legal entities are mentioned in separate segments. These segments are often preceded by known keywords, as shown in figure \ref{fig:case segments}. Once we have identified the beginning and end of each segment, we use them to extract lawyers' names as described in the following subsection.

\subsection{Extraction of lawyers' entities}
To detect lawyers' names throughout the document, we discard, first, all segments except the appellant and the appellee segments. Second, we segment them further into sentences using the sentence tokenizer by Polyglot \footnote{https://polyglot.readthedocs.io/}, which is a Python package providing multilingual natural language processing tools. Third, we only keep sentences containing honorifics used for lawyers such as "Me," "Ma\^itre," or lawyers' keywords like: "represent\'e par." We then use a well-established Named Entity Recognition model by Polyglot \cite{al2015polyglotner} to recognize person entities from the remaining sentences. The model uses pretrained word embeddings from Wikipedia \cite{al2013polyglotemb} to classify whether a word is an entity or not based on its sentence. Last, we consider the extracted named entities appearing in the appellant segment as lawyers of the appellant, and the names appearing in the appellee segment as lawyers of the appellee. It should be noted that decisions without any reference to a lawyer on both sides were overlooked.

\subsection{Extraction of the judge decision}
From the initial segmentation, the final decision of the court is to be found in the conclusion segment of the judgement. Concerning judgments from appeal courts, the court will either confirm the first lower court decision (Tribunal judiciaire) or reverse it. However, the court can also partially confirm the judgment. In other words, the court can decide to accept one of the appellant's requests, and therefore change the first decision partially. Empirically, we noticed that certain words are present in certain types of decisions, and after validation from the domain experts, we resorted to a keyword-based solution:
\begin{itemize}
    \item "Confirme", "Rejete", "Irrecevable": keep the first decision (Appellee "wins")
    \item "Infirme", "Rectifier", "R\'eforme"": change the first decision (Appellant "wins")
\end{itemize}
Out of a sample of 5832 cases, 570 conclusions ($\sim$10\%) include at least a keyword representing both outcomes, in which case we keep the outcome that has most keywords. This is a temporary solution that requires refinement in the future. 

\section{Network analysis of Lawyers}
Once the entities' recognition is complete, we extract all the instances (and their function) in every document. Since courts tend to have a limited number of lawyers, judges and court clerks, cases share the same entities. Therefore, all the cases can be considered a big graph where entities interact with each other.
\subsection{Opposing network of lawyers}
\begin{figure}[h]
  \centering
  \includegraphics[width=\linewidth]{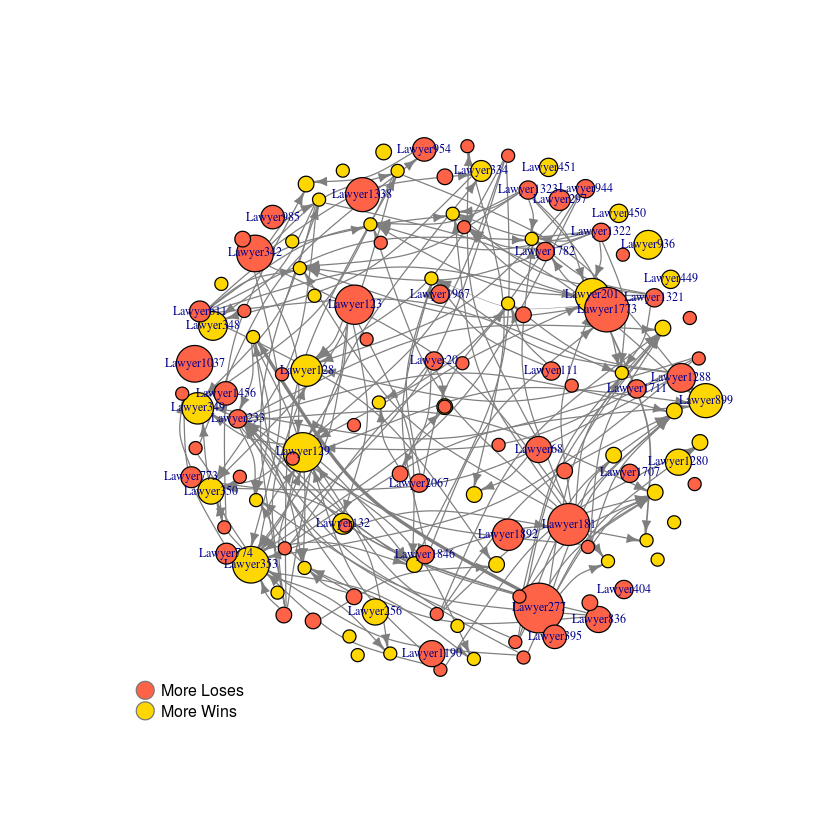}
  \caption{The network of lawyers that have opposed in appeal cases. The width of the edge is a function of the number of wins the source node has over the target. The node color defines whether the lawyer has more losses or more wins, and the size is analogous to total number of cases, large yellow nodes mean the lawyer has won much more cases than he lost and vice versa.}
  \label{fig:opp network}
\end{figure}

We extracted the winning and losing lawyers in each decision. From this, we can define a directed weighted network. We draw an edge between lawyers if they have been opposed. The edge from lawyer i to lawyer j is weighted by the wins $wins_{i,j}$ of lawyer i to lawyer j: $$wins_{i\mapsto j}=aw_{i\mapsto j,acc}+bw_{i\mapsto j,def}$$
Where $w_{i\mapsto j,acc}$ is the number of wins of lawyer i as an appellant and $w_{i\mapsto j,def}$ is the number of wins of lawyer i as an appellee. Parameters 'a' and 'b' are used to weigh more winning as an appellant than winning as an appellee since it is known by legal experts that the event of winning an appeal is less frequent than losing it.We also confirm this intuition by counting the rate of appeals' rejection from our dataset. We get a rejection rate of 0.9.
We collapse both edges between two lawyers into one directed edge weighted by: $|wins_{i\mapsto j} - wins_{j\mapsto i}|\log(wins_{i\mapsto j}+wins_{j\mapsto i}+1)$. In this case the edge direction is determined by the sign of $wins_{i\mapsto j} - wins_{j\mapsto i}$ such that the edge target is the lawyer with most wins. To visualize the most important nodes, we remove lawyers with only one case (899 out of 2146), which leaves us with a network with 1247 nodes and 2182 edges. The resulting network appears in figure \ref{fig:opp network}.
\begin{figure}[h]
  \centering
  \includegraphics[width=0.7\linewidth]{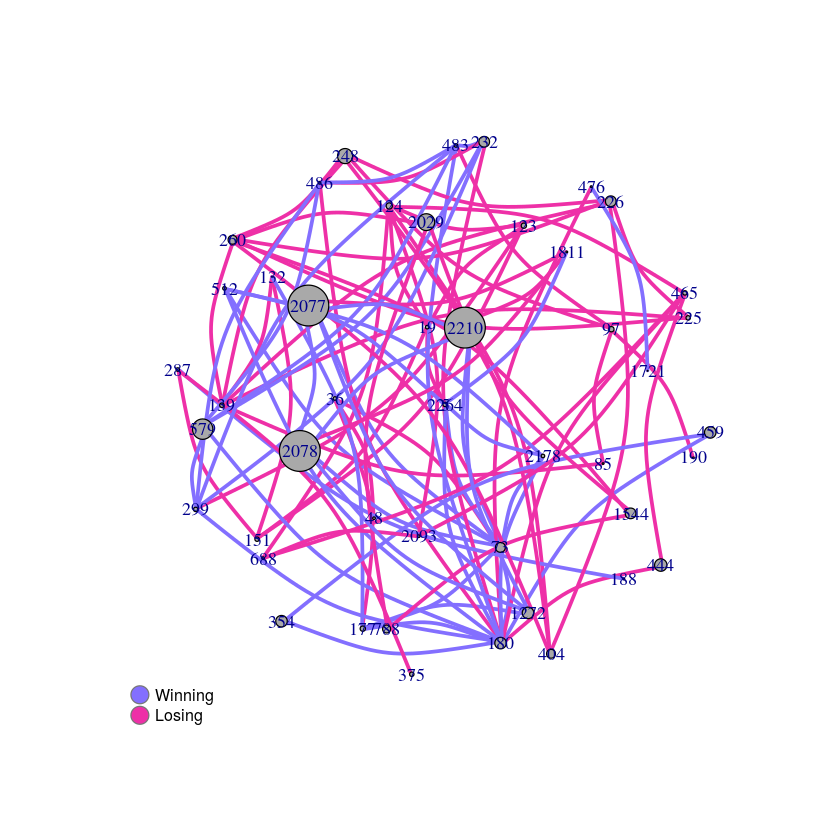}
  \caption{Network of lawyers that have collaborated in court cases. The edge color signifies the sign of the win-loss metric and the width of the absolute value.}
  \label{fig:collab network}
\end{figure}
The edge goes from a "losing" to a "winning" lawyer, and the width of the edge represents the difference in the number of wins. Node size is the number of appeal cases where the lawyer appears and the color captures the win-loss difference.

\subsection{Collaboration network of lawyers}
The collaboration network in figure \ref{fig:collab network} indicates lawyers that have been on the same side during an appeal case. The edges are weighted based on the wins minus the losses, so the network can capture which collaborations are the most successful. We have removed nodes with number of collaborations below a fixed threshold to obtain a decluttered visualization of the network. We obtain a network of 47 nodes and 94 edges out of 2182 nodes and 2950 edges.

\subsection{Lawyers Ranking}
In this section, we suggest three metrics to rank and compare between lawyers. First, we measure the experience of a lawyer by the number of judgments mentioning him as the appellant's or appellee's lawyer. Second, we compute the win-loss rate of lawyers. Third, we calculate the centrality of a lawyer in the opposing network.

\begin{figure}[hbt!]
  \centering
  \begin{subfigure}[t]{0.5\linewidth}
    \centering\includegraphics[width=\textwidth]{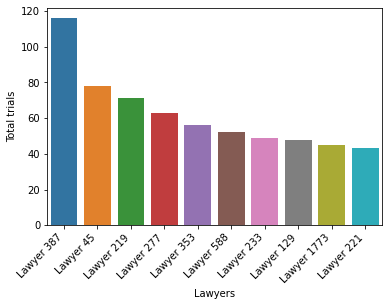}
    \caption{Lawyers ordered by the total number of cases}
    \label{fig: total case}
  \end{subfigure}
  \vfill
  \begin{subfigure}[t]{0.5\linewidth}
    \centering\includegraphics[width=\textwidth]{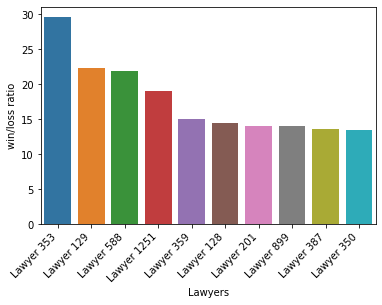}
    \caption{Lawyers ordered by the win/loss ratio}
    \label{fig: win loss ratio}
  \end{subfigure}
  \vfill
  \begin{subfigure}[t]{0.5\linewidth}
    \centering\includegraphics[width=\textwidth]{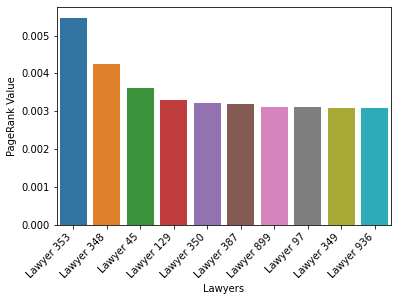}
    \caption{Lawyers ordered by their importance using PageRank algorithm}
    \label{fig: pagerank}
  \end{subfigure}
  \caption{Ranking of lawyers using three different measures}
  \label{fig: lawyers ranking}
\end{figure}

In figure \ref{fig: total case}, lawyers are ranked by their experience in going in front of the court of appeal. However, this measure alone does not indicate the performance of the lawyer. Thus we need to refer to the win/loss ratio to evaluate the performance. Lawyer 353 ranks first in terms of the win/loss ratio instead of fifth in terms of the total number of cases. In this case, lawyer 353 performs better than lawyer 387, who is ranked first in figure \ref{fig: total case} but ranks 9 in terms of win/loss ratio. A weakness of the win/loss ratio ranking is that it does not consider the experience of the opposing lawyer; while the opponent's worth can be a measure of the win's value. To this end, we compute the weighted directed PageRank of the opposing network \ref{fig: pagerank}. As explained in section 4.1, the weights are the number of wins such that wins as an appellant's lawyer counts more than wins as an appellee's lawyer. So edges directed towards lawyers who win more representing an appellant have higher weights than edges directed towards a lawyer who wins more representing an appellee. Therefore top lawyers in figure \ref{fig: pagerank} are lawyers who won against experienced lawyers and who win most as an appellant's lawyer. Lawyer 387 is ranked best than lawyer 350 in terms of win/loss ratio, but worst in terms of PageRank measure. We could explain this difference in the ranking by the fact that the majority of wins of lawyer 350 wins as an appellant's lawyer, while the majority of wins of lawyer 387 wins as an appellee's lawyer. Thus it is recommended for an appellant to choose lawyer 350 rather than lawyer 387.

\subsection{Network analysis of judgments}
In this section, we develop a method to assess cases' difficulty from the perspective of the appellant. More precisely, the aim is to compute the difficulty withing a group of cases dealing with the same legal issues.
First, we built a network of cases to discover communities of cases about similar legal issues. Second, we use the win/loss rate of the appeal as a proxy to its \textit{difficulty}.

\begin{figure}[hbt!]
  \centering
  \begin{subfigure}[t]{0.3\linewidth}
    \centering\includegraphics[width=\textwidth]{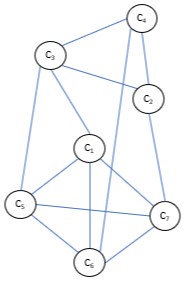}
    \caption{Cases with at least 3 articles in common}
  \end{subfigure}
  \hfill
  \begin{subfigure}[t]{0.3\linewidth}
    \centering\includegraphics[width=\textwidth]{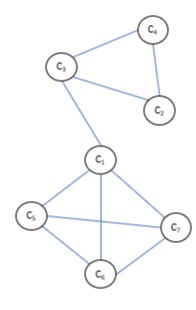}
    \caption{Cases with at least 5 articles in common}
  \end{subfigure}
  \hfill
  \begin{subfigure}[t]{0.3\linewidth}
    \centering\includegraphics[width=\textwidth]{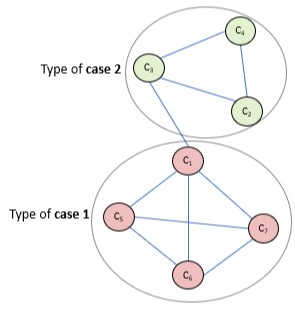}
    \caption{A case graph displaying community structure: two groups of cases with dense internal connections and sparser connections between groups}
  \end{subfigure}
  \caption{Examples of cases graphs and communities for a number of 7 cases}
  \label{fig: case graph ex}
\end{figure}

\begin{figure}[hbt!]
  \centering
  \begin{subfigure}[t]{0.3\linewidth}
    \centering\includegraphics[width=\textwidth]{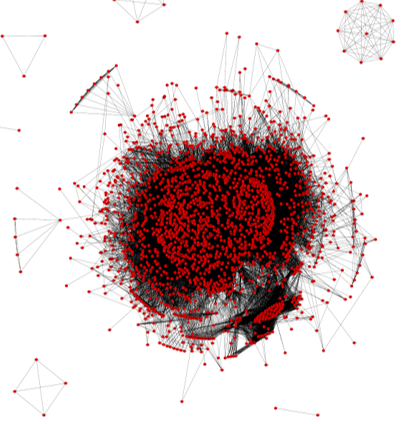}
    \caption{k=2 80,000 edges 1,000 nodes}
  \end{subfigure}
  \hfill
  \begin{subfigure}[t]{0.3\linewidth}
    \centering\includegraphics[width=\textwidth]{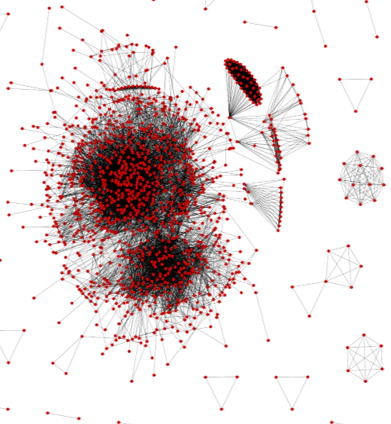}
    \caption{k=3 20,000 edges 600 nodes}
  \end{subfigure}
  \hfill
  \begin{subfigure}[t]{0.3\linewidth}
    \centering\includegraphics[width=\textwidth]{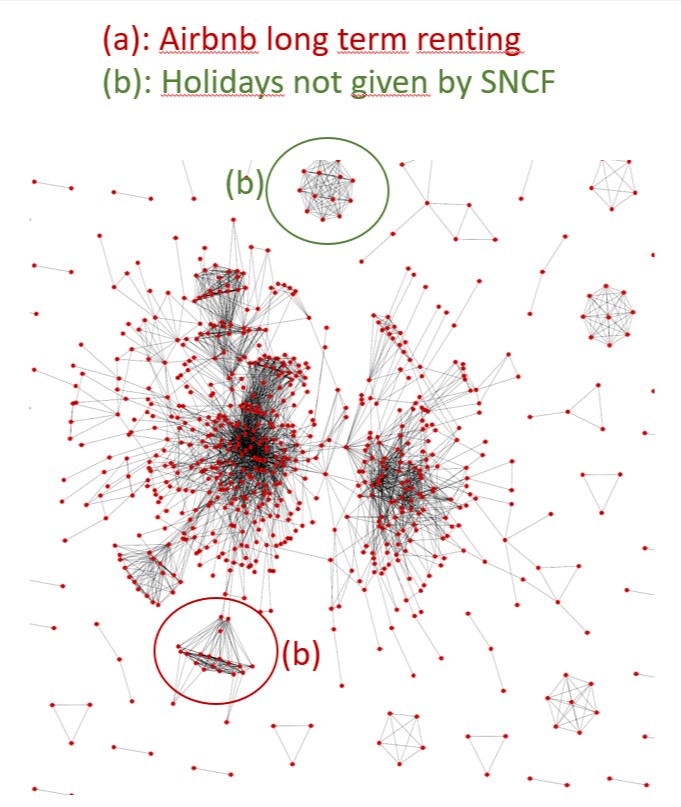}
    \caption{k=4 5,000 edges 400 nodes}
    \label{fig: cases airbnb sncf}
  \end{subfigure}
  \caption{Examples of cases graphs and communities for different values of k, (5500 cases)}
  \label{fig: all cases}
\end{figure}

Graphs encode knowledge and patterns more efficiently \cite{rousseau2013graph, nikolentzos2019message}. The crucial element  is the edges representing some kind of similarity/affiliation among the nodes. Graphs are said to to have the property of community structure \cite{fortunato2010community} when there are groups of vertices with high concentration of internal edges and low concentration of edges between these groups, see example in figure \ref{fig: case graph ex}.c. These special groups are called communities, clusters or modules. In order to build a graph of cases, we needed to connect them with some property that represented similarity. Cases about the same legal issues tend to cite the sames groups of law articles, therefore we define the similarity of two cases by the number of common law articles mentioned in the text of the cases reflecting apparently the thematic similarity among them. Thus we build a network of judgments to discover the communities' structure and natural divisions among the set of studied cases. First, we prepare cases by extracting cited articles of law. We extract articles by using regular expressions. Then we create an edge between two cases if they cite at least k same articles. Figure \ref{fig: all cases} shows graphs of judgments for different values of k. It is evident that as we increase k the graph becomes smaller  with the cases having higher similarity due to the higher number of common articles. 

\begin{figure}[hbt!]
    \centering
    \includegraphics[width=0.8\linewidth]{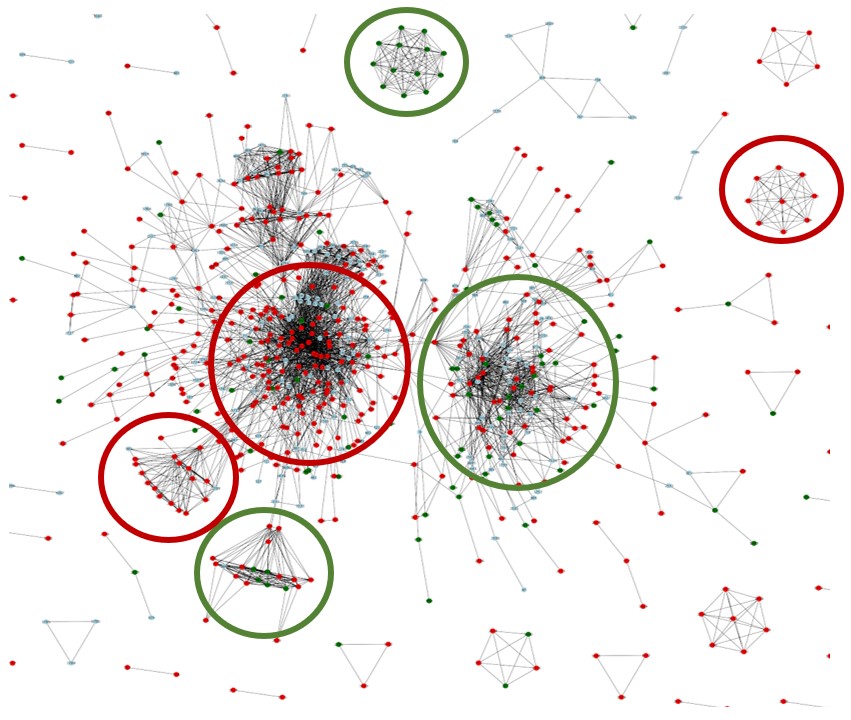}
    \caption{Examples of detected communities. Communities circled in red have a high losing rate. Communities circled in green have a high winning rate.}
    \label{fig:win loss communities}
\end{figure}

We built networks for different values of k from cases of the last three months of 2018, as shown in figure \ref{fig: all cases}. The network naturally groups similar cases in communities. For example, in figure \ref{fig: cases airbnb sncf} cases against the same appellee and about the same issue. 
We also notice, figure \ref{fig:win loss communities} that cases with the same win/loss rate are grouped in the same communities.

\section{Conclusion}
We used NLP methods to extract information from judgments of the French court of appeal. We constructed indicators about the difficulty of lawyers' performance and cases by using network analysis techniques on lawyers' networks and cases' networks. Our objective is to use these indicators to guide laypersons when confronted with the legal systems and contribute to the decrease of the access-to-justice gap by reducing the asymmetry of information characterizing the legal market. The lawyers' ranking could serve to build a system that guides an appellee in choosing a lawyer. However, the lawyers' ranking relies only on wins and losses of lawyers. In future work, we expect to produce a ranking that takes into account the legal area of the case and its difficulty, in such a way that the ranking could be more personalized to the needs of a layperson.



\bibliographystyle{ACM-Reference-Format}
\bibliography{main}


\begin{thebibliography}{27}


\ifx \showCODEN    \undefined \def \showCODEN     #1{\unskip}     \fi
\ifx \showDOI      \undefined \def \showDOI       #1{#1}\fi
\ifx \showISBNx    \undefined \def \showISBNx     #1{\unskip}     \fi
\ifx \showISBNxiii \undefined \def \showISBNxiii  #1{\unskip}     \fi
\ifx \showISSN     \undefined \def \showISSN      #1{\unskip}     \fi
\ifx \showLCCN     \undefined \def \showLCCN      #1{\unskip}     \fi
\ifx \shownote     \undefined \def \shownote      #1{#1}          \fi
\ifx \showarticletitle \undefined \def \showarticletitle #1{#1}   \fi
\ifx \showURL      \undefined \def \showURL       {\relax}        \fi
\providecommand\bibfield[2]{#2}
\providecommand\bibinfo[2]{#2}
\providecommand\natexlab[1]{#1}
\providecommand\showeprint[2][]{arXiv:#2}

\bibitem[\protect\citeauthoryear{??}{sla}{2016}]%
        {slaw_2016}
 \bibinfo{year}{2016}\natexlab{}.
\newblock \showarticletitle{Access to Justice and Market Failure}.
\newblock \bibinfo{journal}{\emph{Slaw}} (\bibinfo{date}{November}
  \bibinfo{year}{2016}).
\newblock
\urldef\tempurl%
\url{http://www.slaw.ca/2016/11/01/access-to-justice-and-market-failure/}
\showURL{%
\tempurl}


\bibitem[\protect\citeauthoryear{??}{oec}{2016}]%
        {oecd2016a2j}
 \bibinfo{year}{2016}\natexlab{}.
\newblock \bibinfo{booktitle}{\emph{Understanding Effective Access to
  Justice}}.
\newblock
\urldef\tempurl%
\url{http://www.oecd.org/gov/Understanding-effective-access-justice-workshop-paper-final.pdf}
\showURL{%
Retrieved April 4, 2020r from \tempurl}


\bibitem[\protect\citeauthoryear{Al-Rfou, Kulkarni, Perozzi, and
  Skiena}{Al-Rfou et~al\mbox{.}}{2015}]%
        {al2015polyglotner}
\bibfield{author}{\bibinfo{person}{Rami Al-Rfou}, \bibinfo{person}{Vivek
  Kulkarni}, \bibinfo{person}{Bryan Perozzi}, {and} \bibinfo{person}{Steven
  Skiena}.} \bibinfo{year}{2015}\natexlab{}.
\newblock \showarticletitle{Polyglot-NER: Massive multilingual named entity
  recognition}. In \bibinfo{booktitle}{\emph{Proceedings of the 2015 SIAM
  International Conference on Data Mining}}. SIAM, \bibinfo{pages}{586--594}.
\newblock


\bibitem[\protect\citeauthoryear{Al-Rfou, Perozzi, and Skiena}{Al-Rfou
  et~al\mbox{.}}{2013}]%
        {al2013polyglotemb}
\bibfield{author}{\bibinfo{person}{Rami Al-Rfou}, \bibinfo{person}{Bryan
  Perozzi}, {and} \bibinfo{person}{Steven Skiena}.}
  \bibinfo{year}{2013}\natexlab{}.
\newblock \showarticletitle{Polyglot: Distributed word representations for
  multilingual nlp}.
\newblock \bibinfo{journal}{\emph{arXiv preprint arXiv:1307.1662}}
  (\bibinfo{year}{2013}).
\newblock


\bibitem[\protect\citeauthoryear{Barriere and Fouret}{Barriere and
  Fouret}{2019}]%
        {barriere2019may}
\bibfield{author}{\bibinfo{person}{Valentin Barriere} {and}
  \bibinfo{person}{Amaury Fouret}.} \bibinfo{year}{2019}\natexlab{}.
\newblock \showarticletitle{May I Check Again?--A simple but efficient way to
  generate and use contextual dictionaries for Named Entity Recognition.
  Application to French Legal Texts}.
\newblock \bibinfo{journal}{\emph{arXiv preprint arXiv:1909.03453}}
  (\bibinfo{year}{2019}).
\newblock


\bibitem[\protect\citeauthoryear{Branting, Weiss, Brown, Pfeifer, Chakraborty,
  Ferro, Pfaff, and Yeh}{Branting et~al\mbox{.}}{2019}]%
        {branting2019semi}
\bibfield{author}{\bibinfo{person}{Karl Branting}, \bibinfo{person}{B Weiss},
  \bibinfo{person}{B Brown}, \bibinfo{person}{C Pfeifer}, \bibinfo{person}{A
  Chakraborty}, \bibinfo{person}{L Ferro}, \bibinfo{person}{M Pfaff}, {and}
  \bibinfo{person}{A Yeh}.} \bibinfo{year}{2019}\natexlab{}.
\newblock \showarticletitle{Semi-Supervised Methods for Explainable Legal
  Prediction}. In \bibinfo{booktitle}{\emph{Proceedings of the Seventeenth
  International Conference on Artificial Intelligence and Law}}.
  \bibinfo{pages}{22--31}.
\newblock


\bibitem[\protect\citeauthoryear{Derl{\'e}n and Lindholm}{Derl{\'e}n and
  Lindholm}{2014}]%
        {derlen2014goodbye}
\bibfield{author}{\bibinfo{person}{Mattias Derl{\'e}n} {and}
  \bibinfo{person}{Johan Lindholm}.} \bibinfo{year}{2014}\natexlab{}.
\newblock \showarticletitle{Goodbye van G end en L oos, Hello B osman? Using
  Network Analysis to Measure the Importance of Individual CJEU Judgments}.
\newblock \bibinfo{journal}{\emph{European Law Journal}} \bibinfo{volume}{20},
  \bibinfo{number}{5} (\bibinfo{year}{2014}), \bibinfo{pages}{667--687}.
\newblock


\bibitem[\protect\citeauthoryear{Epstein, Landes, and Posner}{Epstein
  et~al\mbox{.}}{2013}]%
        {epstein2013behavior}
\bibfield{author}{\bibinfo{person}{Lee Epstein}, \bibinfo{person}{William~M
  Landes}, {and} \bibinfo{person}{Richard~A Posner}.}
  \bibinfo{year}{2013}\natexlab{}.
\newblock \bibinfo{booktitle}{\emph{The behavior of federal judges: a
  theoretical and empirical study of rational choice}}.
\newblock \bibinfo{publisher}{Harvard University Press}.
\newblock


\bibitem[\protect\citeauthoryear{Fortunato}{Fortunato}{2010}]%
        {fortunato2010community}
\bibfield{author}{\bibinfo{person}{Santo Fortunato}.}
  \bibinfo{year}{2010}\natexlab{}.
\newblock \showarticletitle{Community detection in graphs}.
\newblock \bibinfo{journal}{\emph{Physics reports}} \bibinfo{volume}{486},
  \bibinfo{number}{3-5} (\bibinfo{year}{2010}), \bibinfo{pages}{75--174}.
\newblock


\bibitem[\protect\citeauthoryear{Fowler, Johnson, Spriggs, Jeon, and
  Wahlbeck}{Fowler et~al\mbox{.}}{2007}]%
        {fowler2007network}
\bibfield{author}{\bibinfo{person}{James~H Fowler}, \bibinfo{person}{Timothy~R
  Johnson}, \bibinfo{person}{James~F Spriggs}, \bibinfo{person}{Sangick Jeon},
  {and} \bibinfo{person}{Paul~J Wahlbeck}.} \bibinfo{year}{2007}\natexlab{}.
\newblock \showarticletitle{Network analysis and the law: Measuring the legal
  importance of precedents at the US Supreme Court}.
\newblock \bibinfo{journal}{\emph{Political Analysis}} \bibinfo{volume}{15},
  \bibinfo{number}{3} (\bibinfo{year}{2007}), \bibinfo{pages}{324--346}.
\newblock


\bibitem[\protect\citeauthoryear{Greacen, Johnson, and Morris}{Greacen
  et~al\mbox{.}}{2014}]%
        {greacen2014market}
\bibfield{author}{\bibinfo{person}{John~M Greacen}, \bibinfo{person}{Amy~Dunn
  Johnson}, {and} \bibinfo{person}{Vincent Morris}.}
  \bibinfo{year}{2014}\natexlab{}.
\newblock \showarticletitle{From market failure to 100\% access: Toward a civil
  justice continuum}.
\newblock \bibinfo{journal}{\emph{UALR L. Rev.}}  \bibinfo{volume}{37}
  (\bibinfo{year}{2014}), \bibinfo{pages}{551}.
\newblock


\bibitem[\protect\citeauthoryear{Howe, Khang, and Chai}{Howe
  et~al\mbox{.}}{2019}]%
        {howe2019legal}
\bibfield{author}{\bibinfo{person}{Jerrold Soh~Tsin Howe},
  \bibinfo{person}{Lim~How Khang}, {and} \bibinfo{person}{Ian~Ernst Chai}.}
  \bibinfo{year}{2019}\natexlab{}.
\newblock \showarticletitle{Legal Area Classification: A Comparative Study of
  Text Classifiers on Singapore Supreme Court Judgments}.
\newblock \bibinfo{journal}{\emph{arXiv preprint arXiv:1904.06470}}
  (\bibinfo{year}{2019}).
\newblock


\bibitem[\protect\citeauthoryear{Jaro}{Jaro}{1989}]%
        {jaro1989advances}
\bibfield{author}{\bibinfo{person}{Matthew~A Jaro}.}
  \bibinfo{year}{1989}\natexlab{}.
\newblock \showarticletitle{Advances in record-linkage methodology as applied
  to matching the 1985 census of Tampa, Florida}.
\newblock \bibinfo{journal}{\emph{J. Amer. Statist. Assoc.}}
  \bibinfo{volume}{84}, \bibinfo{number}{406} (\bibinfo{year}{1989}),
  \bibinfo{pages}{414--420}.
\newblock


\bibitem[\protect\citeauthoryear{Katz, Bommarito, Michael, and Blackman}{Katz
  et~al\mbox{.}}{2014}]%
        {katz2014predicting}
\bibfield{author}{\bibinfo{person}{Daniel~Martin Katz}, \bibinfo{person}{II
  Bommarito}, \bibinfo{person}{J Michael}, {and} \bibinfo{person}{Josh
  Blackman}.} \bibinfo{year}{2014}\natexlab{}.
\newblock \showarticletitle{Predicting the behavior of the supreme court of the
  united states: A general approach}.
\newblock \bibinfo{journal}{\emph{arXiv preprint arXiv:1407.6333}}
  (\bibinfo{year}{2014}).
\newblock


\bibitem[\protect\citeauthoryear{Kort}{Kort}{1957}]%
        {kort1957predicting}
\bibfield{author}{\bibinfo{person}{Fred Kort}.}
  \bibinfo{year}{1957}\natexlab{}.
\newblock \showarticletitle{Predicting Supreme Court decisions mathematically:
  A quantitative analysis of the “right to counsel” cases}.
\newblock \bibinfo{journal}{\emph{American Political Science Review}}
  \bibinfo{volume}{51}, \bibinfo{number}{1} (\bibinfo{year}{1957}),
  \bibinfo{pages}{1--12}.
\newblock


\bibitem[\protect\citeauthoryear{Long, Tu, Liu, and Sun}{Long
  et~al\mbox{.}}{2019}]%
        {long2019automatic}
\bibfield{author}{\bibinfo{person}{Shangbang Long}, \bibinfo{person}{Cunchao
  Tu}, \bibinfo{person}{Zhiyuan Liu}, {and} \bibinfo{person}{Maosong Sun}.}
  \bibinfo{year}{2019}\natexlab{}.
\newblock \showarticletitle{Automatic judgment prediction via legal reading
  comprehension}. In \bibinfo{booktitle}{\emph{China National Conference on
  Chinese Computational Linguistics}}. Springer, \bibinfo{pages}{558--572}.
\newblock


\bibitem[\protect\citeauthoryear{Michalopoulos, Jacob, and
  Coviello}{Michalopoulos et~al\mbox{.}}{2019}]%
        {michalopoulos2019ai}
\bibfield{author}{\bibinfo{person}{Dennis~P Michalopoulos},
  \bibinfo{person}{Jessica Jacob}, {and} \bibinfo{person}{Alfredo Coviello}.}
  \bibinfo{year}{2019}\natexlab{}.
\newblock \showarticletitle{AI-Enabled Litigation Evaluation: Data-Driven
  Empowerment for Legal Decision Makers}. In
  \bibinfo{booktitle}{\emph{Proceedings of the Seventeenth International
  Conference on Artificial Intelligence and Law}}. \bibinfo{pages}{264--265}.
\newblock


\bibitem[\protect\citeauthoryear{Mok and Mok}{Mok and Mok}{2019}]%
        {mok2019legal}
\bibfield{author}{\bibinfo{person}{Wai~Yin Mok} {and}
  \bibinfo{person}{Jonathan~R Mok}.} \bibinfo{year}{2019}\natexlab{}.
\newblock \showarticletitle{Legal Machine-Learning Analysis: First Steps
  towards AI Assisted Legal Research}. In \bibinfo{booktitle}{\emph{Proceedings
  of the Seventeenth International Conference on Artificial Intelligence and
  Law}}. \bibinfo{pages}{266--267}.
\newblock


\bibitem[\protect\citeauthoryear{Nikolentzos, Tixier, and
  Vazirgiannis}{Nikolentzos et~al\mbox{.}}{2019}]%
        {nikolentzos2019message}
\bibfield{author}{\bibinfo{person}{Giannis Nikolentzos},
  \bibinfo{person}{Antoine J-P Tixier}, {and} \bibinfo{person}{Michalis
  Vazirgiannis}.} \bibinfo{year}{2019}\natexlab{}.
\newblock \showarticletitle{Message Passing Attention Networks for Document
  Understanding}.
\newblock \bibinfo{journal}{\emph{arXiv preprint arXiv:1908.06267}}
  (\bibinfo{year}{2019}).
\newblock


\bibitem[\protect\citeauthoryear{Rachlinski and Wistrich}{Rachlinski and
  Wistrich}{2017}]%
        {rachlinski2017judging}
\bibfield{author}{\bibinfo{person}{Jeffrey~J Rachlinski} {and}
  \bibinfo{person}{Andrew~J Wistrich}.} \bibinfo{year}{2017}\natexlab{}.
\newblock \showarticletitle{Judging the judiciary by the numbers: Empirical
  research on judges}.
\newblock \bibinfo{journal}{\emph{Annual Review of Law and Social Science}}
  \bibinfo{volume}{13} (\bibinfo{year}{2017}), \bibinfo{pages}{203--229}.
\newblock


\bibitem[\protect\citeauthoryear{Rehm, Schneider, Gracia, Revenko, Mireles,
  Khvalchik, Kernerman, Lagzdins, Pinnis, Vasilevskis, et~al\mbox{.}}{Rehm
  et~al\mbox{.}}{2019}]%
        {rehm2019developing}
\bibfield{author}{\bibinfo{person}{Georg Rehm}, \bibinfo{person}{Julian~Moreno
  Schneider}, \bibinfo{person}{Jorge Gracia}, \bibinfo{person}{Artem Revenko},
  \bibinfo{person}{Victor Mireles}, \bibinfo{person}{Maria Khvalchik},
  \bibinfo{person}{Ilan Kernerman}, \bibinfo{person}{Andis Lagzdins},
  \bibinfo{person}{M{\=a}rcis Pinnis}, \bibinfo{person}{Artus Vasilevskis},
  {et~al\mbox{.}}} \bibinfo{year}{2019}\natexlab{}.
\newblock \showarticletitle{Developing and orchestrating a portfolio of natural
  legal language processing and document curation services}. In
  \bibinfo{booktitle}{\emph{Proceedings of the Natural Legal Language
  Processing Workshop 2019}}. \bibinfo{pages}{55--66}.
\newblock


\bibitem[\protect\citeauthoryear{Rousseau and Vazirgiannis}{Rousseau and
  Vazirgiannis}{2013}]%
        {rousseau2013graph}
\bibfield{author}{\bibinfo{person}{Fran{\c{c}}ois Rousseau} {and}
  \bibinfo{person}{Michalis Vazirgiannis}.} \bibinfo{year}{2013}\natexlab{}.
\newblock \showarticletitle{Graph-of-word and TW-IDF: new approach to ad hoc
  IR}. In \bibinfo{booktitle}{\emph{Proceedings of the 22nd ACM international
  conference on Information \& Knowledge Management}}. \bibinfo{pages}{59--68}.
\newblock


\bibitem[\protect\citeauthoryear{Ruhl and Katz}{Ruhl and Katz}{2015}]%
        {ruhl2015measuring}
\bibfield{author}{\bibinfo{person}{JB Ruhl} {and}
  \bibinfo{person}{Daniel~Martin Katz}.} \bibinfo{year}{2015}\natexlab{}.
\newblock \showarticletitle{Measuring, monitoring, and managing legal
  complexity}.
\newblock \bibinfo{journal}{\emph{Iowa L. Rev.}}  \bibinfo{volume}{101}
  (\bibinfo{year}{2015}), \bibinfo{pages}{223}.
\newblock


\bibitem[\protect\citeauthoryear{Sanchez}{Sanchez}{2019}]%
        {sanchez2019sentence}
\bibfield{author}{\bibinfo{person}{George Sanchez}.}
  \bibinfo{year}{2019}\natexlab{}.
\newblock \showarticletitle{Sentence boundary detection in legal text}. In
  \bibinfo{booktitle}{\emph{Proceedings of the Natural Legal Language
  Processing Workshop 2019}}. \bibinfo{pages}{31--38}.
\newblock


\bibitem[\protect\citeauthoryear{Sulea, Zampieri, Vela, and Van~Genabith}{Sulea
  et~al\mbox{.}}{2017}]%
        {sulea2017predicting}
\bibfield{author}{\bibinfo{person}{Octavia-Maria Sulea},
  \bibinfo{person}{Marcos Zampieri}, \bibinfo{person}{Mihaela Vela}, {and}
  \bibinfo{person}{Josef Van~Genabith}.} \bibinfo{year}{2017}\natexlab{}.
\newblock \showarticletitle{Predicting the law area and decisions of french
  supreme court cases}.
\newblock \bibinfo{journal}{\emph{arXiv preprint arXiv:1708.01681}}
  (\bibinfo{year}{2017}).
\newblock


\bibitem[\protect\citeauthoryear{Tarissan and Nollez-Goldbach}{Tarissan and
  Nollez-Goldbach}{2016}]%
        {tarissan2016analysing}
\bibfield{author}{\bibinfo{person}{Fabien Tarissan} {and}
  \bibinfo{person}{Rapha{\"e}lle Nollez-Goldbach}.}
  \bibinfo{year}{2016}\natexlab{}.
\newblock \showarticletitle{Analysing the first case of the international
  criminal court from a network-science perspective}.
\newblock \bibinfo{journal}{\emph{Journal of Complex Networks}}
  \bibinfo{volume}{4}, \bibinfo{number}{4} (\bibinfo{year}{2016}),
  \bibinfo{pages}{616--634}.
\newblock


\bibitem[\protect\citeauthoryear{Vacek, Teo, Song, Nugent, Cowling, and
  Schilder}{Vacek et~al\mbox{.}}{2019}]%
        {vacek2019litigation}
\bibfield{author}{\bibinfo{person}{Thomas Vacek}, \bibinfo{person}{Ronald Teo},
  \bibinfo{person}{Dezhao Song}, \bibinfo{person}{Timothy Nugent},
  \bibinfo{person}{Conner Cowling}, {and} \bibinfo{person}{Frank Schilder}.}
  \bibinfo{year}{2019}\natexlab{}.
\newblock \showarticletitle{Litigation Analytics: Case outcomes extracted from
  US federal court dockets}. In \bibinfo{booktitle}{\emph{Proceedings of the
  Natural Legal Language Processing Workshop 2019}}. \bibinfo{pages}{45--54}.
\newblock


\end{thebibliography}
\clearpage

\end{document}